\documentclass{article}

\usepackage{arxiv}

\usepackage[utf8]{inputenc} 
\usepackage[T1]{fontenc}    
\usepackage{hyperref}       
\usepackage{url}            
\usepackage{booktabs}       
\usepackage{amsfonts}       
\usepackage{nicefrac}       
\usepackage{microtype}      
\usepackage{lipsum}

\usepackage{hyperref}
\usepackage{cleveref}

\usepackage{graphicx}
 \usepackage{algorithm}
\usepackage{algorithmic}

\usepackage{calrsfs}
\DeclareMathAlphabet{\pazocal}{OMS}{zplm}{m}{n}
\newcommand{\La}{\pazocal{L}}
\newcommand{\Ea}{\pazocal{E}}
\newcommand{\Fa}{\pazocal{F}}
\newcommand{\Aa}{\pazocal{A}}
\newcommand{\Ra}{\pazocal{R}}

\newcommand{\Ga}{\pazocal{G}}

\newcommand{\Ta}{\pazocal{T}}

\newcommand{\Ca}{\pazocal{C}}
\newcommand{\Pa}{\pazocal{P}}

\newcommand{\Ba}{\pazocal{B}}

\usepackage{amssymb,amsmath}

\title{Dealing with Incompatibilities among Procedural Goals under Uncertainty}

\author{
  Mariela Morveli-Espinoza \\
  Graduate Program in Electrical and Computer Engineering (CPGEI),\\
 Federal University of Technology - Paran\'{a} (UTFPR),
 Curitiba - Brazil\\
  \texttt{morveli.espinoza@gmail.com} \\
   \And
   Juan Carlos Nieves \\
   Department of Computing Science, Ume{\aa}  University,\\
     Ume{\aa}  - Sweden\\
 \texttt{jcnieves@cs.umu.se}\\
   \And
   Ayslan Trevizan Possebom \\
   Federal Institute of Parana\\
     Paranavai - Brazil\\
 \texttt{possebom@gmail.com}\\
   \And
   Cesar Augusto Tacla \\
  Graduate Program in Electrical and Computer Engineering (CPGEI),\\
 Federal University of Technology - Paran\'{a} (UTFPR),
 Curitiba - Brazil\\
   \texttt{tacla@utfpr.edu.br} \\
}

\newtheorem{definition}{Definition}
\newtheorem{property}{Property}
\newtheorem{example}{Example}
\newtheorem{proof}{Proof}
\newtheorem{theorem}{Theorem}

\begin{document}

\maketitle

\begin{abstract}
By considering rational agents, we focus on the problem of selecting goals out of a set of incompatible ones. We consider three forms of incompatibility introduced by Castelfranchi and Paglieri, namely the terminal, the instrumental (or based on resources), and the superfluity. We represent the agent's plans by means of structured arguments whose premises are pervaded with uncertainty. We measure the strength of these arguments in order to determine the set of compatible goals. We propose two novel ways for calculating the strength of these arguments, depending on the kind of incompatibility that
exists between them. The first one is the logical strength value, it is denoted by a three-dimensional vector, which is calculated from a probabilistic interval associated with each argument. The vector represents the precision of the interval, the location of it, and the combination of precision and location. This type of representation and treatment of the strength of a structured argument has not been defined before by the state of the art. The second way for calculating the strength of the argument is based on the cost of the plans (regarding the necessary resources) and the preference of the goals associated with the plans. Considering our novel approach for measuring the strength of structured arguments, we propose a semantics for the selection of plans and goals that is based on Dung's abstract argumentation theory. Finally, we make a theoretical evaluation of our proposal.

\end{abstract}

\keywords{Argumentation \and Goals Selection \and Uncertainty \and Arguments Strength \and Goals Conflicts}

\section{Introduction}
An intelligent agent may in general pursue multiple procedural goals at the same time\footnote{{\scriptsize A goal is procedural when there is a set of plans for achieving it. These goals are also known as achievement goals \cite{braubach2004goal}.}}. In this situation, some conflicts between goals could arise, in the sense that it is not possible to pursue them   simultaneously. Reasons for not pursuing some goals simultaneously are generally related to the fact that plans for reaching such goals may block each other. Consider the well-known ``cleaner world'' scenario, where a set of robots have the task of cleaning the dirt of an environment. 
Although the main goal of the robots is to clean the environment, during the execution of this task they may pursue some other goals. Furthermore, consider that there exist uncertainties in both actions and sensing.

According to Castelfranchi and Paglieiri \cite{castelfranchi2007role}, at least three forms of incompatibility could emerge:

\begin{itemize}
\item \textit{Terminal incompatibility}: Suppose that at a given moment one of the robots -- let us call him $\mathtt{BOB}$ -- has a technical defect; hence, $\mathtt{BOB}$ begins to pursue the goal ``\textit{going to the workshop to be fixed}''. Recall that $\mathtt{BOB}$ is already pursuing the goal of cleaning the environment; however, if $\mathtt{BOB}$ wants to be fixed he has to stop cleaning. Hence, $\mathtt{BOB}$ cannot pursue both goals at the same time because the plans adopted for each goal lead to an inconsistency, since he needs to be operative to continue cleaning and non-operative to be fixed.

\item  \textit{Instrumental or resource incompatibility}: It arises because the agents have limited resources. Suppose that $\mathtt{BOB}$ is in slot (2,4) and detects two dirty slots, slot (4,2) and slot (4,8). Therefore, he begins to pursue two goals: (i) cleaning slot (4,2) and (ii) cleaning slot (4,8); however, he only has battery for executing the plan of one of the goals. Consequently a conflict due to resource battery arises, and $\mathtt{BOB}$ has to choose which slot to clean.


\item  \textit{Superfluity}: It occurs when the agent pursues two goals that lead to the same end. Suppose that $\mathtt{BOB}$ is in slot (2,2) and he detects dirt in slot (4,5), since it is far from its localization, he has no certainty about the kind of dirt and he begins to pursue the goal ``\textit{cleaning slot (4,5)}''. Another cleaner robot \hbox{-- $\mathtt{TOM}$ --} also detects the same dirty slot and he has the certainty that it is liquid dirt; however, $\mathtt{TOM}$'s battery is quite low, whereby he sends a message to $\mathtt{BOB}$ to mop slot (4,5). Thus, $\mathtt{BOB}$ begins to pursue the goal ``\textit{mopping slot (4,5)}''. It is easy to notice that both goals have the same end, which is that slot (4,5) to be cleaned. 
\end{itemize}

Argumentation is an appropriate approach for reasoning with inconsistent information \cite{dung1995acceptability}. The process of argumentation is based on the construction and the comparison of arguments (considering the so-called attacks among them). Argumentation has been applied for practical reasoning for the generation of desires and plans (e.g., \cite{amgoud2003formal}\cite{amgoud2008constrained}\cite{hulstijn2004combining}\cite{rahwan2006argumentation}). In \cite{amgoud2008constrained} and \cite{rahwan2006argumentation}, the authors represent the agent's plans by means of arguments (these arguments are called instrumental arguments) and the conflicts between plans are expressed in form of attacks. 

In our example, we can have an argument $A$ representing a plan $p$, an argument $B$ representing another plan $p'$, and both attack each other. The question is: what argument will be selected? According to \cite{amgoud2009using}, one can measure the strength of the arguments to refine the notion of acceptability (selection) of arguments. Thus, each argument is measured and a strength value is assigned to it. Then, the arguments' strengths determine the preference of one of them.

In \cite{amgoud2008constrained}, \cite{hulstijn2004combining}, \cite{morveli2019argumentation}, and \cite{rahwan2006argumentation}, the authors use instrumental arguments to represents plans and define possible attacks; however, the agent's beliefs are not pervaded with uncertainty. Besides, in \cite{amgoud2008constrained}, \cite{hulstijn2004combining}, and \cite{rahwan2006argumentation} the actions are not taken into account in the structure of the arguments. The strength of an instrumental argument is measured in \cite{amgoud2008constrained} based on the worth of the goals that make it up and the cost of the plan with respect to the resources it needs to be achieved. 

Other related works are focused on dealing with uncertainty in structured arguments. Hunter \cite{hunter2013probabilistic} assign a probability distribution over the models of the language, which are used to give a probability distribution to the arguments that are constructed using classical logic. Haenni et al. \cite{haenni2009probabilistic} also applies the approach of assigning a probability distribution over the models models of logical arguments. In \cite{amgoud2004using}, Amgoud and Prade propose a possibilistic logic framework where arguments are built from an uncertain knowledge base and a set of prioritized goals. In \cite{nieves2007modality}, Nieves et al. present an argumentation approach based on the ASP's language for encoding knowledge under imprecise or uncertain information. \hbox{Ches\~{n}evar et al. \cite{chesnevar2004logic}} present P-DeLP (Possibilistic Defeasible Logic Programming), which combines features of argumentation theory and logic programming and incorporates a possibilistic treatment of uncertainty. This work was extended in \cite{alsinet2008formalizing} by applying also fuzzy techniques. Thus, the authors use PGL+, a possibilistic logic over G\"{o}del logic extended with fuzzy constants. Schweimeier and Schroeder \cite{schweimeier2004fuzzy} also use fuzzy logic and propose an argumentation framework with fuzzy unification to handle uncertainty. Nevertheless, all these works are not related to goals conflicts and do not represent plans by means of arguments.

Against this background, the aim of this article is to study and propose a way of measuring the strength of instrumental arguments whose premises are pervaded of uncertainty. This will lead us to determine the set of non-conflicting plans and non-conflicting goals the agent can continue pursuing. Thus, the research questions that are addressed in this article are: 

\begin{enumerate}

\item How to measure the strength of an instrumental argument considering that its premises have uncertain elements?, and
\item Given that we use instrumental arguments to determine the incompatibilities between goals, how the uncertainty of the elements of instrumental arguments impact on determining the set of compatible goals? 

\end{enumerate}

In addressing the first question, we use a coherence-based probability logic approach \cite{pfeifer2009framing}. We assign and/or calculate a probabilistic interval for each element of the argument and the interval of the argument is calculated based on the uncertainty of its premises. Lastly, the argument's strength is calculated from this interval. The reason to choose an interval is that it can be used to represent a range of possible values and/or the aggregation of several perceptions. For example, suppose that the dirty sensor of a cleaner robot has a sensitivity error for distinguishing between a dirty slot and a stain on the slot when he is far from the observed slot. In this case, it would be better to represent the probability that such slot is dirty by using an interval. 
Regarding the second question, we use Dung's argumentation semantics in order to obtain the set of compatible goals. 
Thus, the main contributions of this article are:

\begin{itemize}
\item A way of measuring the strength of structured arguments whose premises are pervaded with uncertainty, 
 \item Two different ways of measuring the strength of instrumental arguments: (i) considering their logical structure and (ii) considering the necessary resources.

\item A three-dimensional representation of the logical strength, which allows the agent to compare the argument in more than one way. To the best of our knowledge, the suggested three-dimensional strength of arguments is the first one in its kind for leading with arguments that are pervaded with uncertainty, and

\item A way of selecting goals based on abstract argumentation semantics. 

\end{itemize}

This article has also a practical contribution since it can be applied to real engineering problems. As it can be seen in the example, this kind of approach can be used in robotic applications (e.g., \cite{davids2002urban}\cite{emmi2014new}\cite{tuna2014autonomous}) in order to endow a robot with a system that allows him to recognize and decide about the goals he should pursue. Another possible application is in the spatial planning problem, which aims to rearrange the spatial environment in order to meet the needs of a \hbox{society \cite{ligtenberg2004design}.} As space is a limited resource, it causes that the planner finds conflicts in the desires and expectations about the spatial environment. These desires and expectation can be modeled as a set of restrictions and conditions, which can be considered as goals (e.g., suitability, dependency, and compatibility)\cite{behzadi2013introducing}. Thus, the planner can be seen as a software agent that has to decide among a set of conflicting goals. Although these conflicting goals are not goals the agent wants to achieve, as in the case of the robot, the agent may use this approach in order to resolve the problem and suggest a possible arrangement of the spatial environment. It could also be applied during a design process, in which inconsistencies among design objectives may arise and this results in design conflicts \cite{canbaz2014preventing}. In this case, agents may represent designers that share knowledge and have conflicting interests. This results in a distributed design system that can be simulated as an Multi-Agent System \cite{chira2005multi}\cite{fan2008development}. This last type of application involves more than one agent; however, since there is a conflict among goals, it can be resolved by applying the proposed approach. Another interesting application is in the medication adherence problem. In \cite{ingesonsmart}, the authors propose a coach intelligent agent that is in charge of supporting the medical management of patients. This agent has autonomous reasoning capabilities that allow him to deal with long-term goals in the settings of medication plans.



The rest of the article is organized as follows. Next section presents some necessary technical background related to probabilistic logic. In Section \ref{bloques}, the main building blocks on which this approach is based are defined. Section \ref{ataques} is devoted to the kinds of attacks that may occur between arguments. In Section \ref{fuerza}, we study and present the strength calculation proposal. Section \ref{frames} is focused on the definition of the argumentation framework and on studying how to determine the set of compatible goals by means of argumentation semantics. We present the evaluation of our proposal in terms of fulfilling the postulates of rationality in Section \ref{evalua}. Finally, the conclusions and future work are presented in Section \ref{conclus}.

\section{Probabilistic background}
\label{back}
In this section some necessary technical background is presented. It is based on probabilistic logic inference in the settings of \cite{kern2004combining} and \cite{pfeifer2006inference}.

Let $\La$ be a propositional vocabulary that contains a finite set of propositional symbols. $\wedge$ and $\neg$ denote the logical connectives conjunction and negation. An \textit{event} is defined as follows. The propositional constants \textit{false} and \textit{true}, denoted by $\perp$ and $\top$, respectively, are events. An atomic formula or \textit{atom} is an event. If $\phi$ and $\psi$ are events, then also $\neg \phi$ and $(\phi \wedge \psi)$. A \textit{conditional event} is an expression of the form $\psi | \phi$ and a \textit{conditional constraint} is an expression of the form $(\psi | \phi) [l,u]$ where $l,u\in [0,1]$ are real numbers. The event $\psi$ is called the consequent (or head) and the event $\phi$ its antecedent (or body). \textit{Probabilistic formulas} are defined as follows. Every conditional constraint is a probabilistic formula. If $F$ and $G$ are probabilistic formulas then also $\neg F$ and $(F \wedge G)$. 

One can distinguish between classical and purely probabilistic constraints. \textit{Classical conditional constraints} are of the kind $(\psi | \phi) [1,1]$ or $(\psi | \phi) [0,0]$, while \textit{purely probabilistic conditional constraints} are of the form $(\psi | \phi) [l,u]$ with $l <  1$ and $u > 0$. 

An event $\phi$ is \textit{conjunctive} iff $\phi$ is either $\top$ or a conjunction of atoms. A \textit{conditional event} $\psi | \phi$ is conjunctive (respectively, \textit{1-conjunctive}) iff $\psi$ is a conjunction of atoms (respectively, an atom) and $\phi$ is conjunctive. A conditional constraint $(\psi | \phi) [l,u]$ is conjunctive (respectively, 1-conjunctive) iff $\psi | \phi$ is conjunctive (respectively, 1-conjunctive).

Conjunctive conditional constraints $(\psi | \phi)[l,u]$ with $l \leq u $ are also called \textit{probabilistic Horn clauses}, from which can be defined \textit{probabilistic facts} and \textit{probabilistic rules}, which are of the form $(\psi | \top) [l,u] $ and $(\psi | \phi) [l,u] $, respectively, where $\phi \neq \top$.

We use the coherence-based probability logic to propagate the uncertainty of the premises to the conclusion, more specifically, we use probabilistic $\mathtt{MODUS}$ $\mathtt{PONENS}$. We denote the probabilistic closure $\mathtt{MODUS\; PONENS}$ inference by $\vdash_P$. Finally, the calculation of the conclusion interval is given by \cite{pfeifer2006inference}:

\begin{center}
 $\{(\psi|\phi) [l,u], (\phi|T) [l',u']\} \vdash_P (\psi|T) [l*l', 1-l'+u*l']$ 
\end{center}

\section{Basics of the proposal}
\label{bloques}

In this section, we present the main mental states of the agent; and define a class of structured arguments that represent plans.

In this article, the main mental states of an agent are the following finite bases: 

\begin{itemize}
\item $\Ba$ is a finite set beliefs, 
\item $\Aa$ is a finite base of the actions, 
\item $\Ga$ is a finite base of the goals, and 
\item $\Ra es$ is a finite base of the resources of the agent.
\end{itemize}
 Elements of $\Ba$ and $\Aa$ are probabilistic facts and elements of $\Ga$ and $\Ra es$ are atomic formulas. It holds that $\Ba, \Aa,$ $\Ra es$, and $\Ga$ are pairwise disjoint. Let $\Ba^*= \{b | (b | \top)[l,u] \in B \}$ and $\Aa^*= \{a | (a | \top)[l,u] \in A \}$ be the projections sets of $\Ba$ and $\Aa$, respectively. That is, the elements of $\Ba^*$ and $\Aa^*$ are atomic formulas, which have their correspondent probabilistic conditional constraints in $\Ba$ and $\Aa$, respectively. 
Furthermore, the agent is also equipped with a function $\mathtt{PREF:} \Ga \rightarrow [0,1]$, which returns a real value that denotes the preference value of a given goal (0 stands for the null preference value and 1 for the maximum one) and a resource summary structure $\Ra_{sum}\subseteq \Ra es \times \mathbb{R}^+$ where the first component of a pair is a resource and the second one is the available amount of such resource. 
Let $\mathtt{AVAILABLE\_RES}:\break \Ra es \rightarrow \mathbb{R}^+$ be a function that returns the current amount of a given resource.


The agent has also a set of probabilistic plans that allows him to achieve his goals. In order to analize the possible incompatibilities that could arise among them, we express the plans in terms of instrumental arguments. Thus, the basic building block of an instrumental argument is a probabilistic plan rule, which includes, in the premise, a set of beliefs, a set of goals and a set of actions. All these elements are necessary for the plan to be executed and the goal in the conclusion of the rule to be achieved. 

\begin{definition} \textbf{(Probabilistic plan rule)} A probabilistic plan rule is denoted by a probabilistic rule $(\psi | \phi) [l,u]$ such that $\phi = b_1  \wedge ... \wedge b_n \wedge g_1 \wedge ... \wedge g_m \wedge a_1  \wedge ... \wedge a_l$ and $\psi = g$ where $b_i \in \Ba^*$ (for all $1 \leq i \leq n$), $g_j \in \Ga$ (for all $1 \leq j \leq m$), $a_k \in \Aa^*$ (for all $1 \leq k \leq l$), and $g \in \Ga$. In order to avoid cycles, we require that $\psi \neq g_1$ ... $\psi \neq g_m$. Besides, the number of elements of $\phi$ is finite.

\end{definition}
A probabilistic plan rule expresses that if $b_1  \wedge ... \wedge b_n$ are true at a certain degree, $g_1 \wedge ... \wedge g_m$ are achieved at a certain degree, and $a_1  \wedge ... \wedge a_l$ are accurately performed at a certain degree then $g$ is achieved at a certain degree.\footnote{Achievement goals represent a desired state that an agent wants to reach \cite{dastani2011rich}.}. 

Finally, let $\Pa\Ra$ be the base containing the set of probabilistic plan rules. 

\begin{example}\label{ejemprs} Considering the scenario presented in the introduction section, let us introduce some examples of probabilistic plan rules. Suppose that the environment is a square of 4 rows by 4 columns. So, for the environment to be completely clean, all the slots have to be clean. 

First of all, let us present the beliefs, actions, and goals that are part of the premises of the probabilistic plan rules. Beliefs: $has\_refill, \neg full\_trashcan,$ and $solid\_dirt\_1\_3$. Action: $go\_2\_2$. Goals: $be\_oper, clean\_1\_1,\break..., clean\_4\_4,\ clean, sweep\_4\_4,  in\_wshop$ and $be\_fixed$. We use goal $clean$ to refer to the environment as a whole and we use goals $clean\_1\_1,..., clean\_4\_4$ to refer to each slot of the environment. Thus, to achieve the goal $clean$, all the dirty slots have to be cleaned. Below, we present some probabilistic plan rules:\\
- $(clean\; |\;\: be\_oper \wedge clean\_1\_1 \wedge ... \wedge clean\_4\_4 ) [1,1]$\\
- $(sweep\_4\_4\; |\; \neg full\_trashcan \wedge solid\_dirt\_4\_4 \wedge go\_4\_4)  [0.7,0.9]$\\
- \hbox{$(be\_fixed|\neg be\_oper \wedge has\_refill \wedge in\_wshop)\break[1,1]$}
  \end{example}

We use probabilistic facts and probabilistic plan rules to build probabilistic instrumental arguments, which represent complete plans. Like in \cite{rahwan2006argumentation}, we represent this type of argument using a tree structure; however, in our definition the root is made up of a probabilistic plan rule and the leaves are either beliefs or actions. We can consider these last elements as elementary arguments, since they do not generate sub-trees. We can say that this definition of instrumental argument is new in the state of the art.

\begin{definition} \textbf{(Elementary probabilistic argument)} An elementary probabilistic argument is a tuple $\langle H, (\psi|\top)[l,u]\rangle$ where either $(\psi|\top)[l,u] \in \Aa$ or $(\psi|\top)[l,u] \in \Ba$ and $H=\emptyset$.

\end{definition}

Function $\mathtt{CLAIM}$ returns the claim $\psi$ of a given elementary probabilistic argument. Unlike beliefs and actions, the goals that make up the premise of a probabilistic plan rule generate a tree-structure. Considering that plans may have sub-plans, it is natural that arguments may also have sub-arguments.

\begin{definition} \textbf{(Probabilistic instrumental argument, or complete plan)} A probabilistic instrumental argument is a tuple $\langle \Ta, g \rangle$, where 
$\Ta$ is a finite tree such that:
\begin{itemize}
\item The root of the tree is a structure of the form $\langle H, g\: [l_g,u_g]\rangle$ where:
\begin{itemize}
\item $H=(g | b_1  \wedge ... \wedge b_n \wedge g_1 \wedge ... \wedge g_m \wedge a_1  \wedge ... \wedge a_l)[l,u]$,
\item $l_g,u_g \in [0,1]$ are real numbers that represent the upper and lower probabilities of $g$.
\end{itemize}
\item Since $H=(g | b_1  \wedge ... \wedge b_n \wedge g_1 \wedge ... \wedge g_m \wedge a_1  \wedge ... \wedge a_l)[l,u]$, it has exactly $(n+m+l)$ children, such that $\forall b_i$ ($1 \leq i \leq n$) and $\forall a_k  $ ($1 \leq k \leq l$) there exists an elementary probabilistic argument, and $\forall g_j$ ($1 \leq j \leq m$) there exists a probabilistic instrumental argument, we can call these last arguments of sub-arguments.
\item $H, (b_1| \top [l_{b_1},u_{b_1}]), \ldots, (b_n| \top [l_{b_n},u_{b_n}]),  H_{g_1}, ... , H_{g_m}, (a_1| \top [l_{a_1},u_{a_1}]), \ldots, (a_l| \top [l_{a_l},u_{a_l}]) \vdash_P g [l_g, u_g].$
\end{itemize}

\end{definition}

Let $\mathtt{ARG}$ be the set of all arguments\footnote{{\scriptsize Hereafter, for sake of simplicity, we use only argument to refer to a probabilistic instrumental argument.}}  that are associated to the goals in $\Ga$. We assume that each goal has at least one argument. It is also important to mention that there could be more than one argument for a given goal. Function $\mathtt{SUPPORT}(A)$ returns the set of elementary probabilistic arguments, the main root of the argument, and the roots of the sub-arguments of $A$, $\mathtt{CLAIM}(A)$ returns the claim $g$ of $A$, and $\mathtt{SUB}(A)$ returns the set of sub-arguments of $A$.

In order to obtain the probabilistic interval of the claim of an argument, it has to be applied the probabilistic $\mathtt{MODUS\; PONENS}$ from the leaves to the root.  

\begin{figure}[!htb]
	\centering
	\includegraphics[width=0.8\textwidth]{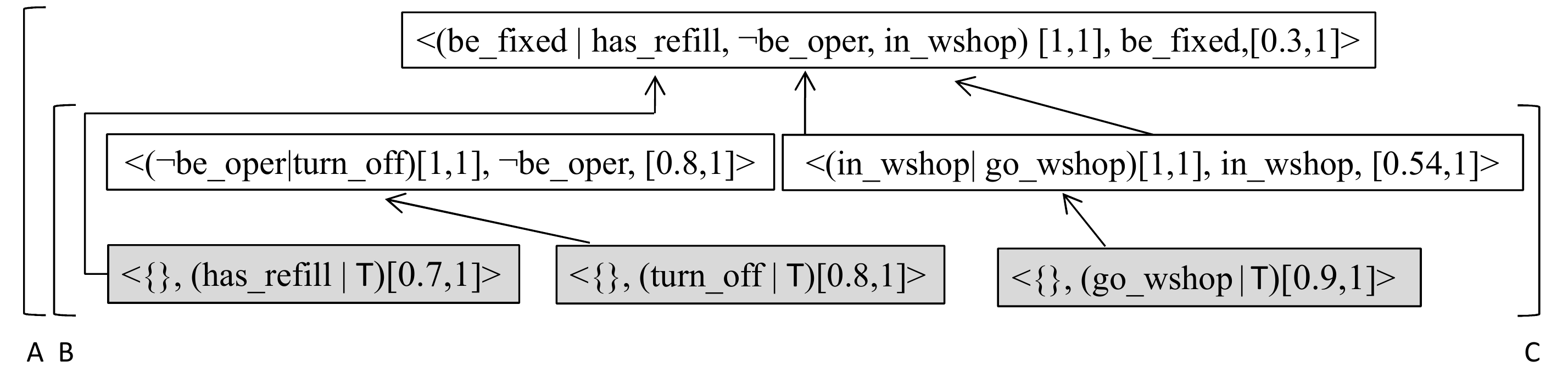} 
	\caption{Tree of probabilistic arguments $A$, $B$, and $C$. Gray-filled rectangles represent the leaves of the tree.}
	\label{argins}
\end{figure}

Figure \ref{argins} shows the tree of the argument $A$ whose claim is goal $be\_fixed$ with the sub-arguments $B$ whose claim is goal $\neg be\_oper$, and $C$ whose claim is goal $in\_wshop$. The values of the probabilistic intervals of the main argument and the sub-arguments are calculated using probabilistic $\mathtt{MODUS\; PONENS}$.

\section{Attacks between arguments}
\label{ataques}

In this section, we focus on the identification of attacks between arguments, which will lead to the identification of incompatibility among goals. The kind of attack depends on the form of incompatibility. The conflicts between arguments are defined over $\mathtt{ARG}$ and are captured by the binary relation $\Ra_x \subseteq \mathtt{ARG} \times \mathtt{ARG}$ (for $x \in \{t,r,s\}$) where each sub-index denotes the form of incompatibility. Thus, $t$ denotes the attack for terminal incompatibility, $r$ the attack for resource incompatibility, and $s$ the attack for superfluity. We denote with $(A, B)$ the attack relation between arguments $A$ and $B$. In other words, if $(A,B) \in \Ra_x$, it means that argument $A$ attacks argument $B$.

\subsection{Terminal incompatibility attack}

We define the terminal incompatibility in terms of attacks among arguments. An argument $A$ attacks another argument $B$ when the claim of any of the sub-arguments of $A$ is the negation of the claim of any of sub-arguments of $B$, and both arguments correspond to plans that allow to achieve different goals.

\begin{definition} \textbf{(Support rebuttal - $\Ra_t$)} Let $A,B \in \mathtt{ARG}$, $[H, \psi] \in \mathtt{SUPPORT}(A)$ and $[H', \psi'] \in \mathtt{SUPPORT}(B)$. We say that $(A,B) \in \Ra_t$ occurs when:

\begin{itemize}
\item $\mathtt{CLAIM}(A) \neq \mathtt{CLAIM}(B)$,
\item $\psi = \neg \psi'$ such that $\psi, \psi' \in \Ba$ or $\psi, \psi' \in \Aa$, or $\psi, \psi' \in \Ga$.

\end{itemize}

\noindent Sub-arguments of arguments that are involved in a support rebuttal are also involved in a support rebuttal between them and with the main arguments. Formally: 

\begin{itemize}
\item If $(A,B) \in \Ra_t$ and $\exists C \in \mathtt{SUB}(B)$ and $\exists D \in \mathtt{SUB}(A)$, then $(C,D) \in \Ra_t$ and $(D,C) \in \Ra_t$. 

\item If $(A,B) \in \Ra_t$ and $\exists C \in \mathtt{SUB}(B)$, then $(A,C) \in \Ra_t$ and $(C,A) \in \Ra_t$.

\end{itemize}

Finally, it holds that $\Ra_t$ is symmetric.
\end{definition}

\begin{figure}[!htb]
	\centering
	\includegraphics[width=0.8\textwidth]{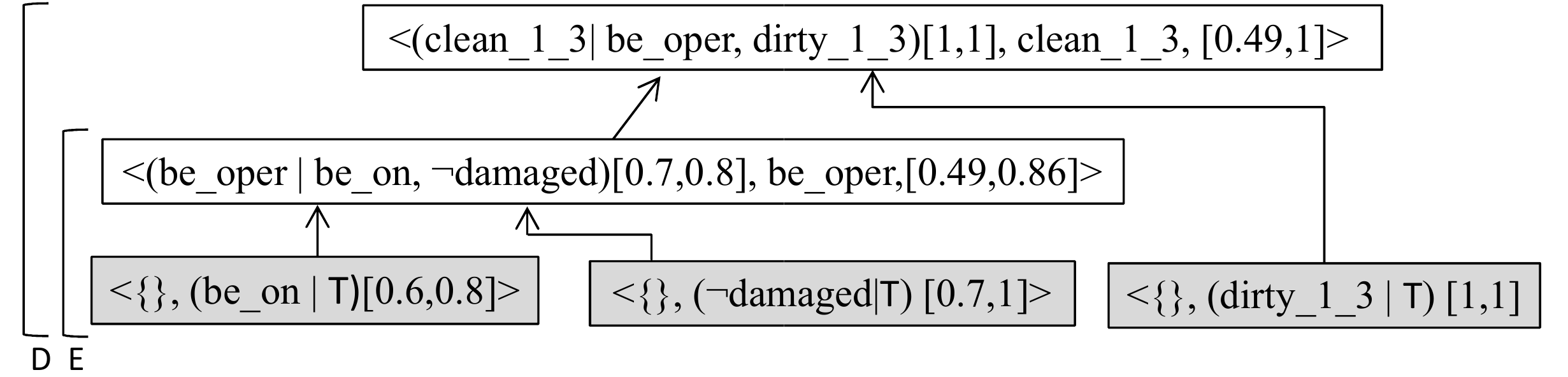} 
	\caption{Probabilistic instrumental argument for goal $clean\_1\_3$. }
	\label{argclean}
\end{figure}

\begin{example} \label{ej-atq} Let $\Ga=\{clean\_1\_3,be\_fixed\}$ be two goals that robot $\mathtt{BOB}$ is currently pursuing. Figure \ref{argins} depicts the arguments $A$, $B$, and $C$. Figure \ref{argclean} shows arguments $D$ whose claim is goal $clean\_1\_3$ and $E$ whose claim is goal $be\_oper$. Since $\mathtt{CLAIM}(D)=\neg \mathtt{CLAIM}(B)$, the following support rebuttals arise: $\{(A,D), (D,A),(A,E),(E,A),(B,D),(D,B), (B,E),(E,B),(C,D),(D,C), (C,E),(E,C)\} \in \Ra_t$.

\end{example}

\subsection{Resources incompatibility attack}

First of all, let us denote with $\mathtt{LIST\_RES\_ARG}(A)$ the list of resources along with their respective amounts that are necessary to perform the plan represented by an argument $A$. We assume that each probabilistic plan rule has a list of the resources that are necessary for its performance. Thus, $\mathtt{LIST\_RES\_ARG}(A)$ is the sum up of the list of every probabilistic plan rule that makes up argument $A$.


Now, we define when a set of instrumental arguments attack each other considering the resources. We evaluate sets of arguments that need the same resource and that are related to plans that allow to achieve different goals. For instance, if three arguments need a certain resource $res$, we compare the current amount of that resource in $\Ra_{sum}$ with the total amount of $res$ the three arguments need. We use function $\mathtt{NEED\_RES}: \mathtt{ARG} \times \Ra es \rightarrow \mathbb{R}^+$ to figure out the amount of a  resource that a given argument needs. Finally recall that function $\mathtt{AVAILABLE\_RES}$ returns the current amount of a given resource.

\begin{definition}\textbf{(Resource attack)} Let $A$ and $B$ be two arguments that need a same resource $res$, such that $A,B \in \mathtt{ARG}$ and $\mathtt{CLAIM}(A) \neq \mathtt{CLAIM}(B)$. We say that $(A, B) \in \Ra_r$  when $\mathtt{NEED\_RES}(A, res) + \mathtt{NEED\_RES}(B, res) > \mathtt{AVAILABLE\_RES}(res)$. It holds that $\Ra_r$ is symmetric.
\end{definition}

\subsection{Superfluity attack}
Superfluity can be defined in terms of the superfluous attack. In this attack, the claims of arguments are evaluated. Thus, an argument $A$ attacks another argument $B$ when they have the same claim. Since each argument may have sub-arguments, this attack is also inherited by the sub-arguments.

\begin{definition}\label{defsuperf} \textbf{(Superfluous attack - $\Ra_s$)} Let $A,B \in \mathtt{ARG}$. We say that $(A,B) \in \Ra_s$ occurs when:

\begin{itemize}

\item $\mathtt{CLAIM}(A) = \mathtt{CLAIM}(B)$,
\item $\mathtt{SUPPORT}(A) \neq \mathtt{SUPPORT}(B)$.

\end{itemize}
Like in the terminal support rebuttal, the sub-arguments of arguments that are involved in a superfluous attack are also involved in a superfluous attack. Finally, it holds that $\Ra_s$ is symmetric.

\end{definition}


\section{Strength calculation}
\label{fuerza}

In this section, the way for calculating the strength of instrumental arguments is presented. We present two different approaches for the strength calculation: (i) considering only the logical structure of the instrumental argument, and (ii) considering the necessary resources.

\subsection{Considering the logical structure}

We base on the approach of Pfeifer \cite{pfeifer2013argument} to calculate the strength of the arguments. This approach uses the values of the probabilistic interval of the claim of the arguments to make the calculation and is based on two criteria: the \textit{precision} of the interval and the \textit{location} of it. Thus, the higher the precision of the interval is and the closer to 1 the location of the interval is, the stronger the argument is. 
We use the notions of precision, location and the combination of both to measure the arguments from different point of views. Then, we next present a three dimensional measure of the arguments strength.

\begin{definition} \textbf{(Logical Strength)} Let $A=\langle \Ta, g \rangle$ be an argument and $\langle H, g[l_g,u_g]\rangle$ be the root of $\Ta$. The logical strength of $A$ --denoted by $\mathtt{STRENGTH}(A)$-- is a three-dimensional vector $\langle \mathtt{CO}(A),\mathtt{PR}(A), \mathtt{LO}(A) \rangle$ where: 

\hspace{4cm}$	      \begin{array}{l}
		 \mathtt{LO}(A)=\left(\frac{l_g+u_g}{2}\right)\\
		 \mathtt{PR}(A)=1-(u_g-l_g)\\
		 \mathtt{CO}(A)=(1-(u_g-l_g)) \times\left(\frac{l_g+u_g}{2}\right)\\
	      \end{array}$

\end{definition}

\begin{example}\label{ex-fuerlog} Consider the arguments of Figure \ref{argins}. The strength values of the main argument and its sub-arguments are: $\mathtt{STRENGTH}(A)=\langle 0.195, 0.3, 0.65 \rangle$, $\mathtt{STRENGTH}(B)=\langle 0.72, 0.8, 0.9 \rangle$, and $\mathtt{STRENGTH}(C)=\langle 0.42, 0.54, 0.77\rangle$.

\end{example}

\subsection{Considering the resources}

For measuring the strength of an argument considering the resources, we take into account the preference value of all the goals involved in the probabilistic instrumental argument, the value of the combination of precision and location of its probabilistic interval, and the cost of achieving the goal in the claim of the argument. Recall that function $\mathtt{NEED\_RES}(A, res)$ returns the amount of resource $res$ that argument $A$ needs. Thus, the cost of an argument $A$ is calculated as follows: $\mathtt{COST}(A)= \sum_{i=1}^{i=n} \mathtt{NEED\_RES}(A, res_i)$, where $n$ is the quantity of resources argument $A$ needs.



\begin{definition} \textbf{(Utility)\footnote{Like in \cite{rahwan2006argumentation}, we call this kind of strength of utility.}} Let $A=\langle \Ta, g \rangle$ be an instrumental argument, the utility strength of $A$ --denoted by $\mathtt{UTILITY}(A)$-- is calculated as follows:

\begin{equation}
\label{equtil}
    \mathtt{UTILITY}(A) =  \sum_{g_i \in \mathtt{SUPPORT}(A) } \mathtt{PREF}(g_i)+\mathtt{COMB}(A)   - \mathtt{COST}(A)   
\end{equation}

\end{definition}

\begin{figure}[!htb]
	\centering
	\includegraphics[width=0.7\textwidth]{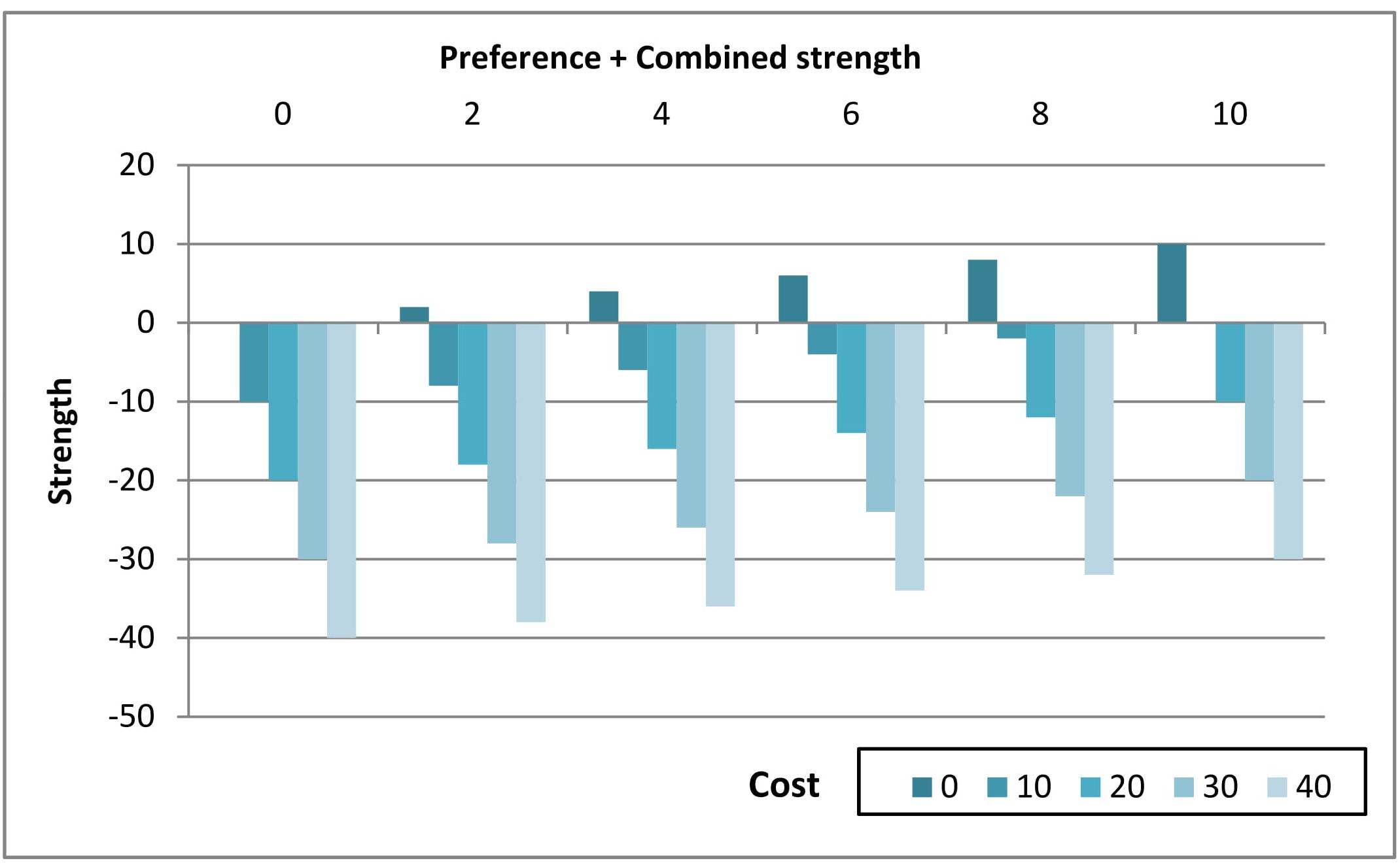} 
	\caption{Example values for demonstrating the behaviour of the utility strength.}
	\label{res-for}
\end{figure}

Notice that we have two positive components and one negative. Figure \ref{res-for} shows the behaviour of the utility strength\footnote{For a graphical representation of the behaviour of the logical strength, the reader is referred to \cite{pfeifer2013argument} pag. 11. }, the $X$ axis denotes the sum of the positive components, the $Y$ axis denotes the utility strength value, and each columns group denotes the preference plus the combined strength value. Thus, in the graphic, we can notice that:

\begin{itemize}
\item Most of the strength values are negative, this is because the values assigned to the preference of the goals and the combined strength of the argument are real values between zero and one, while the values of the cost are positive real numbers with (possibly) values greater than one.
\item In each group of columns, the strongest arguments are displayed with the darkest blue. In the first columns group, this color does not appear because both values are zero.
\item The greater the sum of the preference and the combined strength is and the lower the cost is, the greater the utility strength is. In the figure, the highest value of the strength is given when the sum of the positive components is 10 and the cost is 0. Otherwise, the lowest value of the utility strength occurs when the sum of the positive elements is 0 and the cost is 40. 
\end{itemize}

\subsection{Preference between arguments}

We can now compare two arguments based on these values. This comparison will determine the preference of an argument over another one. Taking into account these three dimensions is specially useful when there is a tie in the value of $\mathtt{CO}(A)$. Consider, for example, that $\mathtt{PR}(A)=0.6$ and $\mathtt{LO}(A)=0.4$, and $\mathtt{PR}(B)=0.4$ and $\mathtt{LO}(B)=0.6$; hence, $\mathtt{CO}(A)=\mathtt{CO}(B)=0.24$. In this case, the agent may determine the strongest argument comparing the other values.

\begin{definition} \textbf{(Preferred argument)} Given two arguments $A$ and $B$. Considering the logical strength, an argument $A$ is more preferred than argument $B$ (denoted by $A \succ B$) if: \\
- $\mathtt{CO}(A)> \mathtt{CO}(B) $, or\\
- $\mathtt{CO}(A)=\mathtt{CO}(B)$ and $ \mathtt{LO}(A)=\mathtt{LO}(B)$ and $ \mathtt{PR}(A)>\mathtt{PR}(B) $, or\\
- $\mathtt{CO}(A)= \mathtt{CO}(B)$ and $ \mathtt{PR}(A)=\mathtt{PR}(B) $ and $ \mathtt{LO}(A)>\mathtt{LO}(B)$. 

\vspace{0.2cm}
\noindent Considering the utility strength, $A$ is more preferred than $B$ (denoted by $A \succ_R B$) if $ \mathtt{UTILITY}(A)> \mathtt{UTILITY}(B)$.

\end{definition}

The election of which value the agent has to compare first (either the precision value or the location one) depends on his interests.

Regarding the logical strength of an argument and the preference relation, the following property shows the relation that exists between the logical strength of a main argument and the logical strength of its sub-arguments.

\begin{property} Let $A=\langle \Ta, g \rangle$ be an argument and $A_1$, ..., $A_n$ all the sub-arguments of $A$. For all $A_i, A_i \succeq A$, where $1 \leq i\leq n$.

\end{property}

In the Example \ref{ex-fuerlog}, one can notice that the logical strength of both sub-arguments is greater than the logical strength of the main argument.

\section{Goals selection}
\label{frames}
In this section, we present an argumentation framework which integrates all the arguments and attacks in a unique framework and will be used to determine the set of compatible goals.

\begin{definition} \textbf{(Argumentation framework)} An argumentation framework is a tuple $\Aa\Fa=\langle \mathtt{ARG}, \Ra \rangle$, where $\Ra= \Ra_t \cup \Ra_s \cup \Ra_r$. 

\end{definition}

Regarding $\Ra$, it could happen that two arguments attack each other in more than one way. For example, suppose that $G$ and $F$ are two arguments such that $(G, F) \in \Ra_t$ and $(G, F) \in \Ra_s$. In these cases, we consider multiple attacks between two arguments as a unique attack in $\Aa\Fa$. 

Hitherto, we have considered that all attacks are symmetrical. However, the strength values of the arguments allow the agent to break such symmetry. Therefore, depending on these values some attacks may be considered successful. Thus, the process of goals selection starts by modifying the attack relation $\Ra$ taking into account the successful attacks.

\begin{definition} \textbf{(Successful attack)}\footnote{In other works, it is called a defeat relation \cite{modgil2014aspic+}.} Let $A,B \in \mathtt{ARG}$ be two arguments, we say that $A$ successfully attacks $B$ when $A \succ B$ or $A \succ_R B$.

\end{definition}

\begin{figure}[!h]
	\centering
	\includegraphics[width=0.8\textwidth]{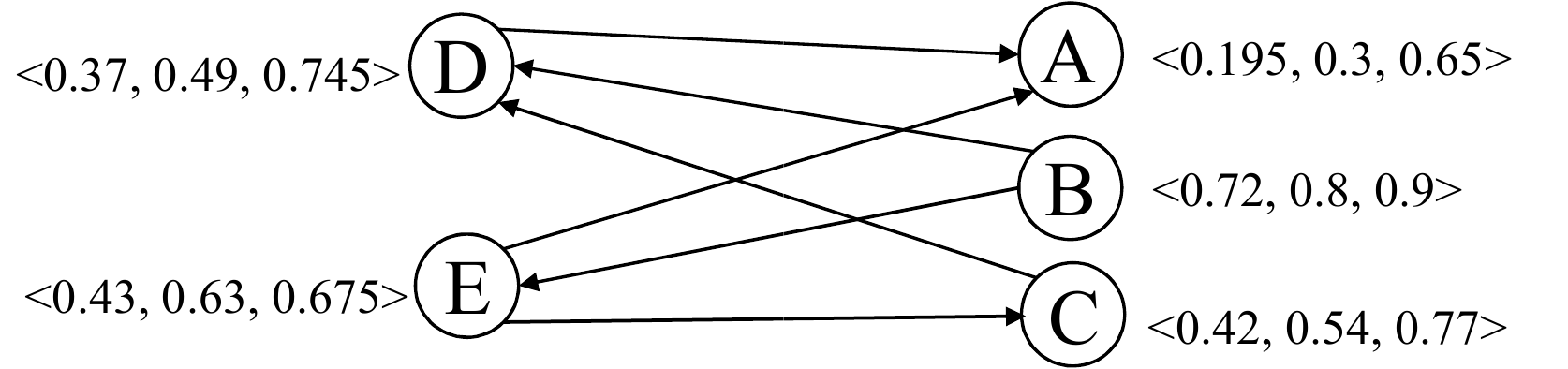} 
	\caption{Graph representation of the argumentation framework $\Aa\Fa$. Nodes represent the arguments and edges the attacks. The strength of each argument is located next to each node.}
	\label{figframe}
\end{figure}

\begin{example} \label{ex-frame} (Cont. Example \ref{ej-atq}) Let $\Aa\Fa'=\langle \{A, B,C,D,E\},\{(D,A),(B,D), (C,D),(E,A),(B,E),(E,C)\} \rangle$ be the AF after considering the successful attack definition. Figure \ref{figframe} shows the graph representation of this framework and the three-dimensional strength of the arguments.
\end{example}

The next step of the selection process is applying an argumentation semantics on the resultant AF, that is, after considering the successful attacks. 

In argumentation theory, acceptability semantics are in charge of returning sets of arguments called extensions which are internally consistent. In order to obtain the set of goals that have no conflicts among them, we will apply the notion of conflict-freeness over the set of arguments to guarantee that no incompatible argument (i.e., plan) is returned by the semantics, and consequently no incompatible goal. Another notion, that we believe is important, is related to the number of compatible goals the agent can continue pursuing; in this way, the idea is to maximize this number. Thus, we propose to apply a semantics based on the notion of conflict-freeness and that also returns those extensions that maximize the number of goals to be pursued.

\begin{definition} \textbf{(Semantics)}\label{basicosarg} Given $\Aa\Fa=\langle \mathtt{ARG}, \Ra' \rangle$ where $\Ra' \subseteq \Ra$ is the modified attack relation after considering the successful attack. Let $\Ea \subseteq \mathtt{ARG}$:
\begin{itemize}
\item $\Ea$ is \textit{conflict-free} if $\forall A, B \in \Ea$, $(A, B) \notin \Ra'$. Let $\Ca\Fa$ be the set of all the conflict-free extensions,
\item $\mathtt{MAX\_GOAL}: \Ca\Fa \rightarrow \Ca\Fa'$, where $\Ca\Fa'=2^{\Ca\Fa}$. This function takes as input a set of conflict-free sets and returns those maximal (w.r.t set inclusion) sets that allow the agent to achieve the greatest number of pursuable goals. Sub-goals are not taken in to account in this function.
\item$\mathtt{MAX\_UTIL}: \Ca\Fa' \rightarrow  2^{\Ca\Fa'}$. This function takes as input the set of conflict-free sets $\Ca\Fa'$ and returns those with the maximum utility for the agent in terms of preference value. The utility of each extension is calculated by summing up the preference value of the main goals of the extension. In this function, sub-goals are not taken in to account either.
\end{itemize}
\end{definition}

The final step of the selection process is to obtain the set of compatible goals from the set of compatible plans. 

\begin{definition} \textbf{(Projection function)} Let $\Ca\Fa''$ be a set of extensions returned by $\mathtt{MAX\_UTIL}$. Function $\mathtt{COMP\_GOALS}: \Ca\Fa'' \rightarrow  2^{\Ga}$ takes as input an extension of $\Ca\Fa'$ and returns the set of compatible goals that are associated to the arguments in the extension.

\end{definition}

Notice that function $\mathtt{COMP\_GOALS}$ is applied to each extension of $\Ca\Fa'$; hence, there could be more than one different set of compatible goals. In such case, the agent has to choose the set of compatible goals he will continue to pursue according to his interests.

\begin{example}(Cont. Example \ref{ex-frame}) After applying the conflict-free semantics the conflict-free extensions are $\Ca\Fa=\{\{\},\{C\}, \{B\}, \{B,C\},\{A\}, \{A,C\}, \{A,B\}, \{A,B,C\}, \{E\}, \{D\}, \{D,E\}\}$. Then we apply\break $\mathtt{MAX\_GOAL}(\Ca\Fa)$ and we obtain: $\Ca\Fa'=\{\{A,B,C\},\{D,E\}\}$. Extension $\{A,B,C\}$ allows the agent to achieve the top-goal $be\_fixed$ whereas extension $\{D,E\}$ allows the agent to achieve the top-goal $clean\_1\_3$. Since there is a tie to break, the agent employs the preference value of the main goal. Assume that $\mathtt{PREF}(\{D,E\})=0.9$ and $\mathtt{PREF}(\{A,B,C\})=0.75$. Consequently,  $\mathtt{MAX\_UTIL}(\{A,B,C\},\{D,E\})=\{D,E\}$.

Finally, we apply the projection function: $\mathtt{COMP\_GOALS}(\{D,E\})=\{clean\_1\_3, be\_oper\}$.

\end{example}
\section{Evaluation of the approach}
\label{evalua}
In this section, we evaluate the proposed approach and prove that it satisfies the rationality postulates proposed in \cite{caminada2007evaluation}. Firstly, let us define the following notation:

\indent$\mathtt{BEL}(\Ea)= \bigcup_{A \in \Ea} (\mathtt{BODY}(\mathtt{SUPPORT}(A)) \cap \Ba)$,\\
\indent$\mathtt{ACT}(\Ea)= \bigcup_{A \in \Ea} (\mathtt{BODY}(\mathtt{SUPPORT}(A)) \cap \Aa)$\\
\indent$\mathtt{GOA}(\Ea)= \bigcup_{A \in \Ea} (\mathtt{BODY}(\mathtt{SUPPORT}(A)) \cap \Ga)$

It is important to understand the following definition before presenting the results because both consistency and closure are specified based on justified conclusions. A goal that is the conclusion of an argument in any extension can be regarded as a justified conclusion, even if it is not in all extensions. From a more restrictive point of view, a goal can be regarded as a justified conclusion when it is the conclusion of an argument that belongs to all the extensions. 

\begin{definition}\textbf{(Justified conclusions)} Let $\Aa\Fa=\langle \mathtt{ARG}, \Ra \rangle$ be a general AF and $\{\Ea_1, ..., \Ea_n\}$ ($n \geq 1$) be its set of extensions under the conflict-free semantics.

\begin{itemize}
\item $\mathtt{CONCS}(\Ea_i)=\{\mathtt{CLAIM}(A) | A \in \Ea_i\} (1 \leq i \leq n )$.
\item $\mathtt{Output}=\bigcap_{i=1,...,n} \mathtt{CONCS}(\Ea_i)$.
\end{itemize}

$\mathtt{CONCS}(\Ea_i)$ denotes the justified conclusions for a given extension $\Ea_i$ and $\mathtt{Output}$ denotes the conclusions that are supported by at least one argument
in each extension.
\end{definition}

An important property required in \cite{caminada2007evaluation} is direct consistency. An argumentation system satisfies direct consistency if its set of justified conclusions and the different sets of conclusions corresponding to each extension are consistent. This property is important in our approach because it guarantees that the agent will only pursue non-conflicting goals. 

\begin{theorem} \label{teorconst}\textbf{(Direct consistency)} Let $\Aa\Fa=\langle \mathtt{ARG}, \Ra\rangle$ be a general AF and $\Ea_1, ..., \Ea_n$ its conflict-free extensions. $\forall \Ea_i, i=1,...,n$, it holds that:

\begin{enumerate}
\item The set of beliefs $\mathtt{BEL}(\Ea_i)$ is a consistent set of literals.
\item The set of actions $\mathtt{ACT}(\Ea_i)$ is a consistent set of literals.
\item The set of goals $\mathtt{GOA}(\Ea_i)$ is a consistent set of literals.
\item The set of goals $\mathtt{GOA}(\Ea_i)$ has no superfluous conflicting goals.
\end{enumerate}

\end{theorem} 

\begin{proof} Let $\Ea_i$ be an extension of $\Aa\Fa$.

1. Let us show that $\mathtt{BEL}(\Ea_i)$ is a consistent set of literals. Suppose that $\mathtt{BEL}(\Ea_i)$ is inconsistent. This means that $\exists b, \neg b \in \mathtt{BEL}(\Ea_i)$. Consider that $\exists\; A, B \in \Ea_i$ such that $[\{\},b] \in \mathtt{SUPPORT}(A)$ and $[\{\},\neg b] \in \mathtt{SUPPORT}(B)$. This means that there is a supports rebuttal between $A$ and $B$ due to $b$ and $\neg b$. Since it contradicts the fact that $\Ea_i$ is conflict-free, we can say that $\mathtt{BEL}(\Ea_i)$ is consistent. 

2. Let us show that $\mathtt{ACT}(\Ea_i)$ is a consistent set of literals. Suppose that $\mathtt{ACT}(\Ea_i)$ is inconsistent. This means that $\exists\;a, \neg a \in \mathtt{ACT}(\Ea_i)$. Consider that $\exists A, B \in \Ea_i$ such that $[\{\},a] \in \mathtt{SUPPORT}(A)$ and $[\{\},\neg a] \in \mathtt{SUPPORT}(B)$. This means that there is a supports rebuttal between $A$ and $B$ due to $a$ and $\neg a$. Since it contradicts the fact that $\Ea_i$ is conflict-free, we can say that $\mathtt{ACT}(\Ea_i)$ is consistent. 

3. Let us show that $\mathtt{GOA}(\Ea_i)$ is a consistent set of literals. Suppose that $\mathtt{GOA}(\Ea_i)$ is inconsistent. This means that $\exists\;g, \neg g \in \mathtt{GOA}(\Ea_i)$. Consider that $\exists\; A, B \in \Ea_i$ such that $[\{\},g] \in \mathtt{SUPPORT}(A)$ and $[\{\},\neg g] \in \mathtt{SUPPORT}(B)$. This means that there is a supports rebuttal between $A$ and $B$ due to $g$ and $\neg g$. Since it contradicts the fact that $\Ea_i$ is conflict-free, we can say that $\mathtt{GOA}(\Ea_i)$ is consistent. 

4. Let us show that $\mathtt{GOA}(\Ea_i)$ has no superfluous conflicting goals. Suppose that $\exists g, g' \in \mathtt{GOA}(\Ea_i)$ such that $g$ and $g'$ are superfluous goals. Since $g$ and $g'$ are superfluous goals, it means that $\exists A, B \in \Ea_i$ such that $\mathtt{CLAIM}(A)=\mathtt{CLAIM}(B)=g=g'$ and $\mathtt{SUPPORT}(A)\neq\mathtt{SUPPORT}(B)$. Hence, there is a superfluous attack between $A$ and $B$. This contradicts the fact that $\Ea_i$ is conflict free.

Finally, since all the extensions obtained from $\Aa\Fa$ are consistent, then $\mathtt{OUTPUT}$ is also consistent.  
\end{proof}

Next property is closure, the idea of closure is that the set of justified conclusions of every extension should be closed under the set of plan rules $\Pa\Ra$. That is, if $g$ is a conclusion of an extension and there exists a plan rule $g \rightarrow g'$, then $g'$ should also be a conclusion of the same extension. Next definition states the closure of the set of plan rules of the agent.

\begin{definition}\label{defclos} (\textbf{Closure of $\Pa\Ra$)} Let $\Fa \subseteq \Ba^* \cup \Aa^* \cup \Ga$. The closure of $\Fa$ under the set $\Pa\Ra$ of plan rules, denoted by $\Ca l_{\Pa\Ra}(\Fa)$, is the smallest set such that:
\begin{itemize}
\item $\Fa \subseteq \Ca l_{\Pa\Ra}(\Fa)$
\item If $(g|b_1 \wedge ...\wedge b_n \wedge g_1 \wedge ...\wedge g_m \wedge a_1, ...\wedge a_l)[l,u] \in \Pa\Ra$ and

 $b_1, ..., b_n, g_1, ..., g_m, a_1, ..., a_l \in \Ca l_{\Pa\Ra}(\Fa)$ then $g \in \Ca l_{\Pa\Ra}(\Fa)$.
\end{itemize}


If $\Fa = \Ca l_{\Pa\Ra}(\Fa)$, then $\Fa$ is said to be closed under the set $\Pa\Ra$.
\end{definition}

In our approach closure is important because it guarantees that all the goals that can be inferred from $\Pa\Ra$ be evaluated in terms of their possible conflicts, which in turn guarantees that the agent only will pursue non-conflicting goals.

\begin{theorem} \textbf{(Closure)} Let $\Pa\Ra$ be a set of probabilistic plan rules, $\Aa\Fa=\langle \mathtt{ARG}, \Ra' \rangle$ be an argumentation framework built from $\Pa\Ra$. $\mathtt{Output}$ is its set of justified conclusions, and $\Ea_1, ..., \Ea_n$ its extensions under the conflict-free semantics. $\Aa\Fa$ satisfies closure iff:

\begin{enumerate} 
\item $\mathtt{CONCS}(\Ea_i)=\Ca l_{\Pa\Ra}(\mathtt{CONCS}(\Ea_i))$ for each $1 \leq i \leq n$.
\item $\mathtt{Output} = \Ca l_{\Pa\Ra}(\mathtt{Output})$.
\end{enumerate}

\end{theorem}

\begin{proof}

Let us call $\mathtt{ARG}_{Cl}$ the arguments that can be built from $\mathtt{BEL}(\Ea) \cup \mathtt{ACT}(\Ea) \cup \mathtt{GOA}(\Ea)$ and $\Pa\Ra$. 

Given that $\mathtt{CONCS}(\Ea_i)=\Ca l_{\Pa\Ra}(\mathtt{CONCS}(\Ea_i))$, we will proof that $\forall \Ea_i, i=1, ..., n$, it holds that $\Ea_i = \mathtt{ARG}_{Cl}$; thus, we will first proof that $\Ea_i \subseteq \mathtt{ARG}_{Cl}$ and then that $\mathtt{ARG}_{Cl} \subseteq \Ea_i$.

\begin{enumerate}
\item $\Ea_i \subseteq \mathtt{ARG}_{Cl}$: This is trivial.
\item $\mathtt{ARG}_{Cl} \subseteq \Ea_i$: Suppose that $\mathtt{ARG}_{Cl} \nsubseteq \Ea_i$; hence, $\exists A \in \mathtt{ARG}_{Cl}$ such that $A \not\in \Ea_i $. Since  $A \not\in \Ea_i $, it means that $\exists B \in \Ea_i$ such that $(B,A) \in \Ra'$ or $(A,B) \in \Ra'$. There are three situation in which this happens, one for each form of incompatibility:

\begin{enumerate}
\item \textit{When the attack is a supports rebuttal} ($(B,A) \in \Ra_t$ or $(A,B) \in \Ra_t$): This means that $\exists [H, \psi] \in \mathtt{SUPPORT}(B)$ and $\exists [H', \psi'] \in \mathtt{SUPPORT}(A)$ such that $\psi = \neg \psi'$, considering that $\psi, \psi' \in \Ba$ or $\psi, \psi' \in \Ga$ or $\psi, \psi' \in \Aa$. Since both $A$ and $B$ are built from $\mathtt{BEL}(\Ea_i)$, $\mathtt{ACT}(\Ea_i)$, and $\mathtt{GOA}(\Ea_i)$ this would mean that $\mathtt{BEL}(\Ea_i)$, $\mathtt{ACT}(\Ea_i)$, or $\mathtt{GOA}(\Ea_i)$ are inconsistent, which contradicts the first, second, and third items of Theorem \ref{teorconst}.

\item \textit{When it is a superfluous attack} ($(B,A) \in \Ra_s$ or $(A,B) \in \Ra_s$): Let $g=\mathtt{CLAIM}(A)$ and $g'=\mathtt{CLAIM}(B)$. Since there is a superfluous attack between $A$ and $B$, it means that $g$ and $g'$ are superfluous goals. Given that both arguments are built from $\mathtt{BEL}(\Ea_i)$, $\mathtt{ACT}(\Ea_i)$, and $\mathtt{GOA}(\Ea_i)$, we can say that their conclusions are also part of $\mathtt{GOA}(\Ea_i)$, i.e. $g,g' \in \mathtt{GOA}(\Ea_i)$. This contradicts the last item of Theorem \ref{teorconst}, which proofs that there is no superfluous conflicting goals in $\mathtt{GOA}(\Ea_i)$.

\end{enumerate}

\end{enumerate}


\end{proof}

The last property our proposal should satisfy is indirect consistency. This property means that (i) the closure under the set of probabilistic plan rules of the set of justified conclusions is consistent, and (ii) for each extension, the closure under the set of probabilistic plan rules of its conclusions is consistent. 

\begin{theorem} \textbf{(Indirect consistency)} Let $\Pa\Ra$ be a set of probabilistic plan rules and $\Aa\Fa=\langle \mathtt{ARG}, \Ra' \rangle$. $\mathtt{OUTPUT}$ is its set of justified conclusions, and $\Ea_1, ..., \Ea_n$ its conflict-free extensions under the conflict-free semantics. $\Aa\Fa=\langle \mathtt{ARG}, \Ra' \rangle$ satisfies indirect consistency iff:

- $\Ca l_{\Pa\Ra}(\mathtt{CONCS}(\Ea_i))$ is consistent for each $1 \leq i \leq n$.\\
\indent - $\Ca l_{\Pa\Ra}$ ($\mathtt{OUTPUT}$) is consistent.

\end{theorem}

\begin{proof}

Based on Proposition 7 defined in \cite{caminada2007evaluation}\footnote{Extracted from \cite{caminada2007evaluation}: ``\textbf{Proposition 7}. Let $\langle A,Def \rangle$ be an argumentation system. If $\langle A,Def \rangle$ satisfies closure and direct consistency, then it also satisfies indirect consistency.''.}, we can say that our proposal satisfies indirect inconsistency since it satisfies closure and direct consistency.

\end{proof}

\section{Conclusions and future work}
\label{conclus}

This article presents: (i) a way for measuring instrumental arguments which components are pervaded of uncertainty and (ii) an argumentation-based approach for selecting compatible goals from a set of incompatible ones. We use abstract argumentation theory because we have noticed that the problem of goals selection can be compared to the problem of calculating an extension in abstract
argumentation. Therefore, we have adapted some concepts of abstract argumentation to our problem. \\

With respect to the research questions presented in the introduction, we can now state the following:

\begin{enumerate}

\item In order to represent the uncertainty of the elements of an argument, we use probabilistic intervals, which express the certainty degree of the beliefs, actions, and goals that made up the argument. Since the structure of an argument is a tree, we apply probabilistic $\mathtt{MODUS\; PONENS}$ from the leaves to the root to obtain the probabilistic interval of the goal that is the claim of the argument. We represent the logical strength of an argument by means of a three-dimensional vector, which includes the values of the precision and the location of the interval of the claim, and the combined value of both precision and location. Besides, we calculate the utility strength of an argument based on the resources that are necessary for performing the plan associated to the arguments. Both types of strength determine the successful attacks in the argumentation framework.
\item The uncertainty of the elements that make up an argument impacts on the final set of compatible goals in the following way: the set of compatible goals depends on the attack relation among the arguments. We consider that an attack is successful when an argument is more (or equal) preferred than other. Since the preference relation is determined using the logical strength, this means that the certainty degree determines if an attack is successful or not.

\end{enumerate}

Comparing with related work, we can state the following:

\begin{itemize}

\item In \cite{amgoud2008constrained}, \cite{hulstijn2004combining}, and \cite{rahwan2006argumentation}, they use arguments to represent plans, determine conflicts, and select compatible plans. The main differences of our proposal with these approaches are: (i) the structure of our arguments includes beliefs and actions, which are not considered in the related works, and (ii) they deal with the conflicts that are similar to the attack defined for terminal incompatibility; however, the superfluous attacks not taken into account.

\item Unlike our proposal, \cite{amgoud2008constrained}, \cite{hulstijn2004combining}, \cite{morveli2019argumentation}, and \cite{rahwan2006argumentation} do not consider uncertainty in the elements of the arguments.

\item Our approach is based on probabilistic logic; however, unlike \cite{hunter2013probabilistic} and \cite{haenni2009probabilistic}, we use an interval representation of the probabilistic value of the components of the arguments.

\item Finally, with respect to the strength calculation of arguments, in \cite{amgoud2008constrained}, it is measured based on the worth of the goals that made up it and the cost of the plan in relation to the resources it needs to be achieved. We calculate the strength based on the probabilistic values and propose a three-dimensional measure that allows to evaluate an argument from different perspectives, which is an advantage of our proposal. 

\end{itemize}

A direct future work of this research is its practical application on the medical adherence problem that is being tackled by researchers of the Department of Computing Science together with the Department of Community Medicine and Rehabilitation of the Umea University. In this project, a Medication Coach Intelligent Agent (MCIA) is being developed \cite{ingesonsmart}. Such agent perceives the environment through a smart augmented reality device --the Microsoft HoloLens-- and has autonomous reasoning capabilities. We believe that the proposed approach fits perfectly in this project because the MCIA has to deal with uncertain information perceived through the HoloLens sensors and uses such information during his reasoning cycle, which includes the selection of goals.

Another future (more theoretical) research direction is related to probability logic. We have used the coherence-based probability logic to propagate the uncertainty from the support to the claim of the argument. We would like to study other approaches of probabilistic logic in order to obtain tighter intervals, if possible. 

Finally, we plan to implement this proposal in Java and want to study how to integrate it with JASON \cite{bordini2005bdi}, a BDI-based platform for developing intelligent agents.

\section*{Acknowledgment}
This work is partially founded by CAPES (Coordena\c{c}\~{a}o de Aperfei\c{c}oamento de Pessoal de N\'{i}vel Superior).

\bibliographystyle{unsrt}  
\bibliography{iberamia1}  


\end{document}